\documentclass[lettersize,journal]{IEEEtran}
\usepackage{amsmath,amsfonts}
\usepackage{algorithmic}
\usepackage{algorithm}
\usepackage{array}
\usepackage[caption=false,font=normalsize,labelfont=sf,textfont=sf]{subfig}
\usepackage{textcomp}
\usepackage{stfloats}
\usepackage{enumitem}
\usepackage{url}
\usepackage{verbatim}
\usepackage{graphicx}
\usepackage{cite}
\usepackage[many]{tcolorbox}
\usepackage[table,xcdraw]{xcolor}
\usepackage{xcolor}
\usepackage{multirow}
\usepackage[table]{xcolor}
\usepackage{caption}
\newtcolorbox{promptbox}[1][]{
    enhanced,
    colback=gray!5,      
    colframe=black!70,   
    boxrule=0.8pt,       
    arc=4pt,             
    left=8pt, right=8pt, top=8pt, bottom=8pt,
    fonttitle=\bfseries,
    title=#1,            
    drop shadow=black!10 
}
\usepackage{hyperref}
\hypersetup{
    colorlinks=false,      
    citebordercolor=0 1 0, 
    linkbordercolor=1 0 0  
}

\begin{document}

\title{MODF-SIR: A Multi-agent Omni-modal Distilled Framework for Social Intelligence Reasoning}

\author{%
	Shang Ma,
	Jisheng Dang,
      Wencan Zhang,
        Yifan Zhang,
            Bimei Wang,

    	Hong Peng,
	Bin Hu,~\IEEEmembership{Fellow, ~IEEE},
    	Qi Tian,~\IEEEmembership{Fellow, ~IEEE},
	Tat-Seng Chua%
    \thanks{This work was supported by the National Natural Science Foundation of China (Grants No. 62227807 and U24B20186). This work was also supported by the Supercomputing Center of Lanzhou University.}
    \thanks{Shang Ma, Jisheng Dang, Yifan Zhang, Bimei Wang, and Hong Peng  are with the School of information Science and Engineering, Lanzhou University, LanZhou, China (E-mail: dangjisheng@lzu.edu.cn).}
\thanks{Bin Hu is with the School of Medical Technology, Beijing Institute of Technology, Beijing 100081, China (E-mail: bh@bit.edu.cn). }

 \thanks{Qi Tian is with Cloud and AI BU, Huawei, Shenzhen, Guangdong 518129,
China (e-mail: tian.qi1@huawei.com).}
    \thanks{Wencan Zhang and Tat-Seng Chua are with the School of Computing, National University of Singapore, Singapore 119077 (E-mail: dcscts@nus.edu.sg). }
    \thanks{*Corresponding author: Jisheng Dang, Bimei Wang and Bin Hu.}
}

\maketitle

\begin{abstract}
We propose a multi-agent collaborative framework built upon a lightweight Multimodal Large Language Model (MLLM), specifically designed for social intelligence reasoning. A key feature of our approach is that both the training and inference phases are augmented via knowledge distillation. Within this architecture, multi-modal data pertinent to social intelligence is precisely localized. Furthermore, relevant long-tail events are identified, extracted, and rendered as formatted, explicit text. This formatting strategy prevents critical long-tail information from being overshadowed by head events and environmental noise during the tokenization process. Specifically, we integrate Test-Time Adaptation (TTA) across the entire reasoning pipeline, encompassing the extraction and representation of long-tail events, Chain-of-Thought (CoT) prompting, and self-reflection. This TTA mechanism is also distillation-enhanced, utilizing Low-Rank Adaptation (LoRA) to fine-tune the foundation model exclusively for instance-level reasoning. Extensive evaluations against various open-source and proprietary AI models across multiple benchmarks demonstrate the effectiveness of the proposed framework. With around \textit{30\%} of training data from IntentTrain, we achieve state-of-the-art results.
Codes are available at \url{https://github.com/eeee-sys/MODF-SIR}, demo is available at \url{https://huggingface.co/spaces/Harry-1234/MODF-SIR}, LoRA is available at \url{https://huggingface.co/Harry-1234/MODF-SIR} and the dataset for training router is available at \url{https://huggingface.co/datasets/Harry-1234/IntentRouterTrain}.
\end{abstract}


\section{Introduction}
\IEEEPARstart{S}{ocial} intelligence denotes the cognitive process by which intelligent systems infer and interpret human intentions, emotions, interpersonal dynamics, and implicit social norms based on multimodal social signals (e.g., spoken language, facial expressions, paralinguistics, and physical actions). It is a fundamental yet challenging problem in artificial intelligence \cite{rabinowitz2018machine}. 
In real-world scenarios, humans rarely communicate through explicit and fully specified instructions. Instead, intentions are implicitly conveyed through a combination of language, visual cues, actions, and social context \cite{tomasello2005understanding}. As a result, intelligent  systems must go beyond surface-level perception and develop the ability to infer latent goals, interpret interpersonal dynamics, and reason about evolving interaction process~\cite{schaafsma2015deconstructing}.
It is important that human intentions are expressed through diverse signals such as speech, facial expressions, body movements, and environmental evidence, which are tightly coupled and dynamically evolving over time \cite{tomasello2005understanding}\cite{frith2006neural}. Effective intention understanding therefore requires integrating perception and action cues and explicitly modeling their interactions, rather than treating them independently \cite{frith2006neural}\cite{yang2021mtag}. 
Beyond multi-modal integration, this problem is deeply rooted in social cognition and theory of mind. Humans infer others’ intentions by reasoning about their beliefs, goals, and internal states, a process supported by both action understanding and higher-level mentalization mechanisms that are dynamically engaged depending on the interaction context. 
Consequently, affective analysis must account not only for individual behavior but also for multi-agent dynamics, shared goals, and contextual dependencies, making it significantly more complex than traditional perception tasks \cite{drijvers2023multimodal}.
\begin{figure}[t]
    \centering
    \includegraphics[trim=1mm 1mm 1mm 1mm, clip, width=\linewidth]{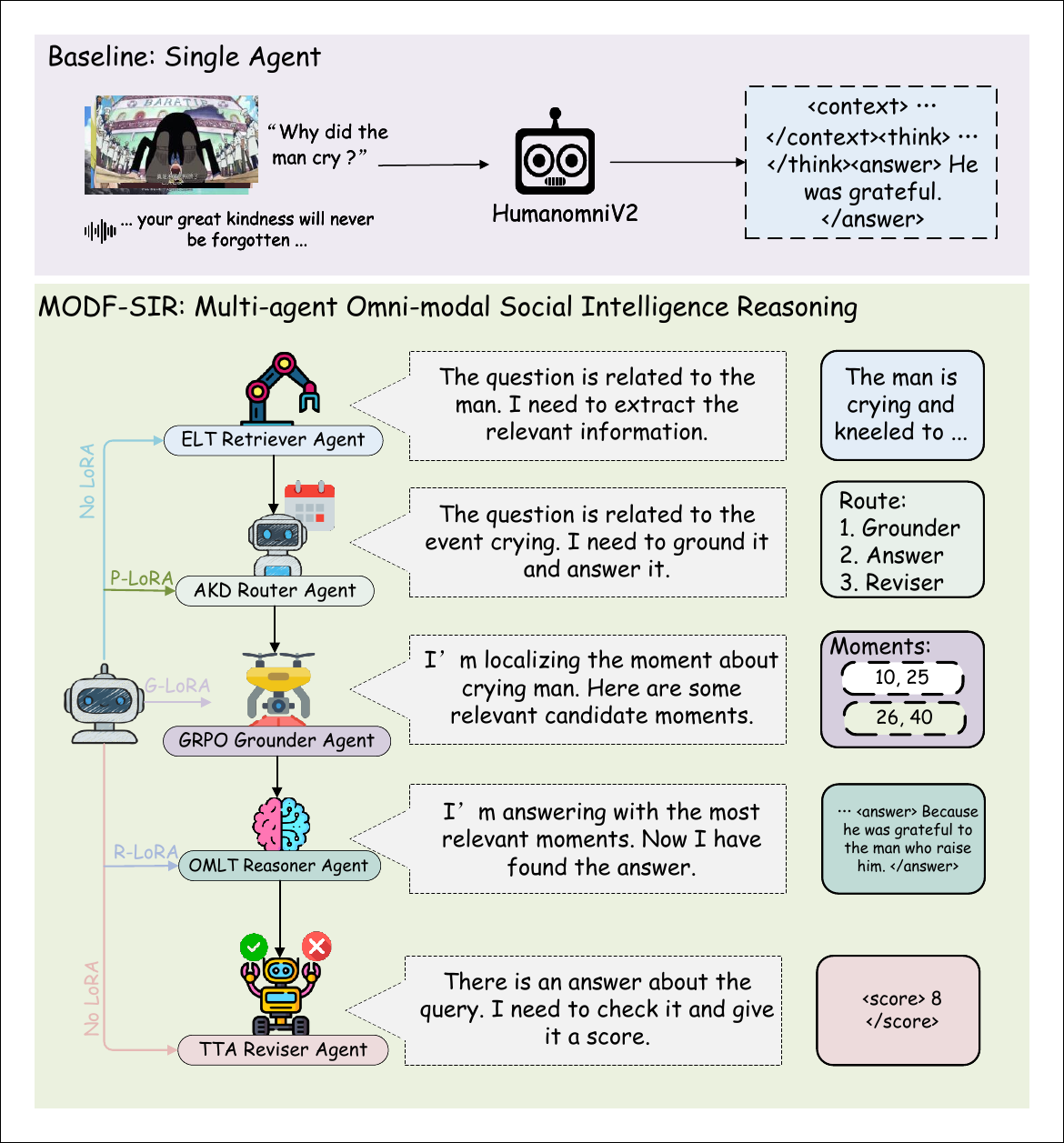}
    \caption{\textbf{The motivation of our method.} Traditional method employs a black-box reasoning paradigm, which may cause problems such as hallucinations. Our method employs multi-agent strategy, visualizing the reasoning steps.}
    \label{motivation}
\end{figure}

To overcome these limitations, in this paper, we propose a lightweight MLLM-based, distillation-augmented, multi-agent collaborative framework for social intelligence reasoning, called a Multi-agent Omni-modal Distilled Framework for Social Intelligence Reasoning (\textbf{MODF-SIR}) with the following features: 

(i) \textbf{MODF-SIR} employs a two-stage retrieval mechanism for omni-modal data. The initial retrieval is executed by a prompt-guided foundation model (ELT Retriever Agent), while the subsequent retrieval is performed by a distillation-enhanced model (OMLT Reasoner Agent). Inspired by the psychological ``Dual-Process Theory'' \cite{kahneman2011thinking}, the first stage acts as a coarse-grained process (System 1) designed to inform the routing decisions of the AKD Router Agent. Conversely, the second stage is a fine-grained process (System 2) that underpins the social intelligence reasoning of the OMLT Reasoner Agent. Crucially, the results from both retrieval stages are serialized into natural language. This textualization explicitly accentuates long-tail events associated with social intelligence, thereby preventing these critical signals from being overshadowed by dominant head events or background noise during the tokenization process. 

(ii) Guided by the outputs from the OMLT Retriever Agent, the AKD Router Agent within \textbf{MODF-SIR} dictates the subsequent reasoning paradigm. It dynamically routes the process toward either standard reasoning or social intelligence reasoning, with the latter specifically targeting long-tail events within the omni-modal data. This conditional routing mechanism ensures the optimal allocation of computational resources. 

(iii) Long-tail events are inherently subtle and transient. Consequently, conducting an exhaustive global search to identify these events across unconstrained omni-modal data would incur a prohibitive computational overhead for \textbf{MODF-SIR}. To alleviate this, the framework introduces the \textbf{GRPO Grounder Agent}, which precisely localizes the data segments most relevant to the user query, thereby significantly narrowing the effective search space.

(iv) The reasoning process within the MODF-SIR framework is structured into three distinct stages, which are collaboratively executed by the OMLT Reasoner Agent and the TTA Reviser Agent. In the initial stage, the OMLT Reasoner Agent executes retrieval on the omni-modal data segments localized by the GRPO Grounder Agent. This step aims to extract query-relevant occurrences, explicitly prioritizing the long-tail events that are indispensable for social intelligence reasoning. Concurrently, the framework incorporates a teacher-guided evaluation loop. An external large model, acting as a teacher, assesses the quality of the retrieved results. If the evaluation indicates suboptimal performance, the base model undergoes dynamic parameter updates via Low-of-Rank Adaptation (LoRA) and iteratively re-executes the retrieval process until the teacher model's criteria are fully satisfied. In the second phase, the OMLT Reasoner Agent leverages the retrieved evidence to perform query-driven Chain-of-Thought (CoT) reasoning. Mirroring the previous stage, the generated reasoning trajectory is evaluated by the external teacher. Suboptimal outputs trigger iterative, LoRA-based parameter updates and CoT regeneration until the teacher's criteria are completely satisfied. Finally, the TTA Reviser Agent evaluates the OMLT Reasoner Agent's output. Suboptimal answers trigger LoRA-based parameter updates and subsequent re-reasoning. This self-correction module explicitly leverages the well-known ``generation-evaluation gap'' \cite{zheng2023judging}, capitalizing on the consensus that LLMs are inherently more proficient at evaluating outputs than generating them.

Like the TTA Reviser Agent, \textbf{MODF-SIR}'s parameter updates during inference constitute Test-Time Adaptation (TTA). Specifically, \textbf{MODF-SIR} applies instance-specific LoRA updates tailored to the current omni-modal input and query. Post-inference, these LoRA weights are discarded, completely restoring the base model. To evaluate the effectiveness of our method, We tested our method on three benchmarks, including Worldsense \cite{hong2025worldsense}, Daily-omni \cite{zhou2025daily}, IntentBench \cite{yang2025humanomniv2}. Leveraging a multi-agent design that jointly produces temporally grounded evidence and accurate responses, our framework demonstrates strong capability in modeling human intent and psychological dynamics. Overall, our contributions are as below:
\begin{itemize}
    \item We propose \textbf{MODF-SIR}, a unified omni-modal reasoning framework that pioneers the application of multi-agent collaboration in the field of social intelligence reasoning. Our framework introduces dynamic strategy selection via a routing agent, enabling the model to adaptively determine whether to perform temporal grounding or direct reasoning based on input complexity.
    \item We introduce \textbf{GRPO Grounder} and \textbf{TTA Reviser}. We train the video locator implemented by the autoregressive method using the GRPO algorithm and fine-tune the reasoning module during testing using the test-time adaption and REINFORCE with Baseline algorithms. This method enables our framework to have sample-level answering capabilities.
    \item We construct a new training data for multi-agent routing by performing knowledge distillation. This distilled data provides high-quality pseudo-labels essential for training the AKD Router Agent, significantly enhancing its decision-making capabilities. With around \textbf{30\%} of training data from IntentTrain \cite{yang2025humanomniv2}, we achieve state-of-the-art results.
    \item \textbf{MODF-SIR} achieves state-of-the-art results across three Benchmarks: IntentBench \cite{yang2025humanomniv2}, Daily-Omni \cite{zhou2025daily}, WorldSense \cite{hong2025worldsense}. Notably, our approach surpasses a host of commercial closed-source and open-source models, including GPT-4o\cite{hurst2024gpt}, Gemini-2.5-Pro (think)\cite{google2025gemini25pro}. Extensive ablations further confirm its effectiveness.
\end{itemize}

\section{Related Work}

\subsection{Temporal Grounding and Reasoning in Videos}
Significant progress in video understanding has enabled a wide range of tasks, including video captioning, video question answering, and video-text retrieval, which primarily focus on semantic-level understanding of visual content \cite{vinyals2016show}\cite{jang2017tgif} However, these approaches often lack fine-grained temporal alignment between queries and visual evidence, limiting their ability to provide precise and interpretable reasoning, especially in long and complex videos.
The task of temporal grounding has addressed this limitation. Existing methods have achieved strong performance in temporal localization. Nevertheless, these models are mainly optimized for boundary prediction and often fail to capture the underlying reasoning process or provide interpretable justifications for their predictions \cite{liu2023survey}. Recent advances in video reasoning and grounded question answering further highlight the need for integrating temporal localization with high-level reasoning. These tasks require models to not only identify relevant segments but also perform multi-step reasoning over temporally distributed evidence.
\subsection{Model Refinement}
Model refinement has emerged as an important paradigm for improving the performance of LLMs beyond standard one-pass inference. Early approaches focus on inference-time refinement, where models iteratively improve their outputs through strategies such as self-consistency, tool-augmented reasoning, and structured search \cite{wang2022self}\cite{shinn2023reflexion}. Another line of work leverages reinforcement learning (RL) to enhance reasoning capabilities by optimizing model behavior through reward signals \cite{ouyang2022training}. Recent advances demonstrate that RL can significantly improve multi-step reasoning and decision-making process. Nevertheless, most of these models rely on vision-related tasks heavily, overlooking the  omni-modality information. In parallel, parameter-efficient fine-tuning techniques, such as loew-rank adaptation, provide a practical way to update model parameters with minimal computational overhead. While effective, these methods are primarily designed for offline adaptation and are rarely integrated into iterative reasoning frameworks for continuous refinement.

\section{method}
\begin{figure*}[htbp]
    \centering
    \includegraphics[trim=1mm 1mm 1mm 1mm, clip, width=\linewidth]{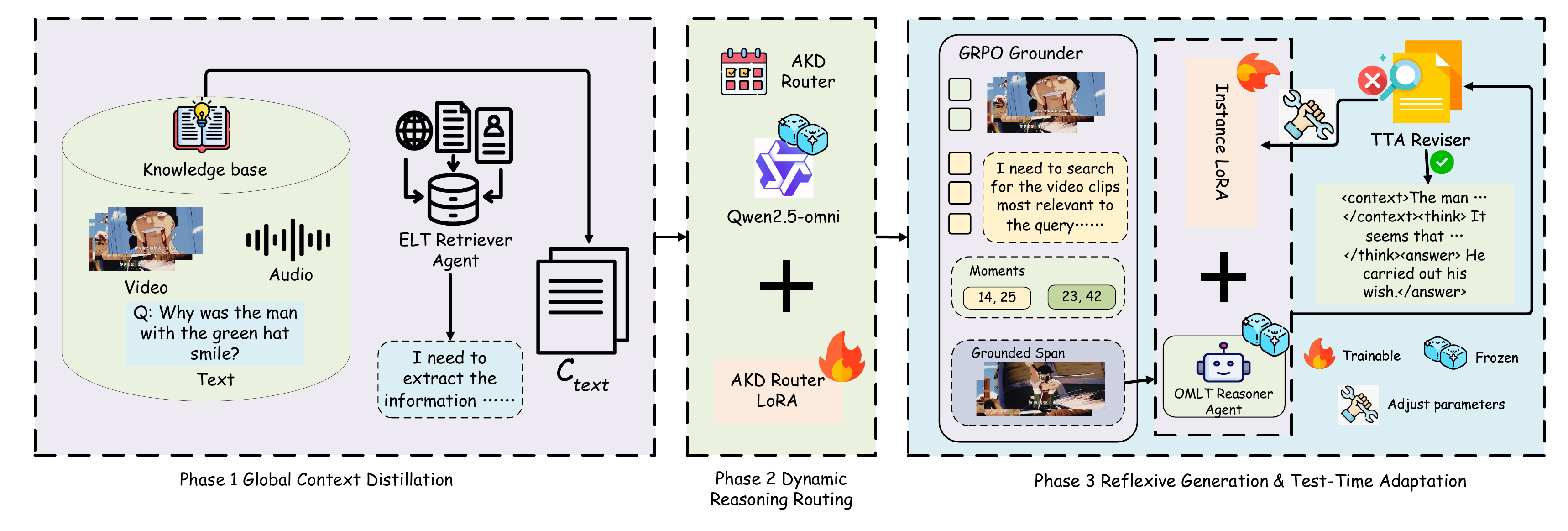}
    \caption{\textbf{The overall workflow of our MODF-SIR.} Given a video, audio and query. MODF-SIR would activate different agents based on different situations and perform step-by-step reasoning.}
    \label{method}
\end{figure*}
\subsection{Framework}
When processing complex real-world human interactions, MLLMs typically rely on a flat, black-box end-to-end reasoning paradigm \cite{song2025modularized}. When confronted with long-duration, information-dense audio-visual streams, this paradigm often suffers from severe ``cognitive overload'' \cite{lian2026verbatim}, impairing the model’s ability to capture implicit human intentions, subtle emotional dynamics, and evolving psychological states. Moreover, the lack of a self-reflection mechanism implies that early hallucinations in spatio-temporal grounding can propagate, leading to irreversible error cascades in downstream reasoning \cite{zhu2025llm} \cite{ji2025denoising}. 

To address these limitations, we draw inspiration from the ``Dual-Process Theory'' \cite{kahneman2011thinking} and propose \textbf{MODF-SIR} (Fig. \ref{method}), a multi-agent collaborative framework designed for Social Intelligent Reasoning. It leverages knowledge distillation across training and inference, while incorporating TTA and Episodic LoRA for instance-level, self-corrective reasoning. MODF-SIR is designed to extract and analyze long-tail multimodal events, such as subtle behavioral or facial shifts, to decode human intent and emotion. Using extended CoT reasoning, it answers implicit user queries regarding emotional and intentional dynamics. The framework consists of the following agents:

\textbf{Endogenous Long-Tail (ELT) Retriever Agent.} The ELT Retriever Agent, a System-1 module \cite{kahneman2011thinking}, rapidly scans multi-modal inputs for long-tail events. It converts these observations into text to provide routing criteria for MODF-SIR's subsequent reasoning.

\textbf{Asymmetric Knowledge Distilled (AKD) Router Agent.} Trained via Asymmetric Knowledge Distillation, the AKD Router Agent utilizes the ELT Retriever Agent's textual event descriptions to route the instance to the appropriate downstream pathway. Additionally, it rewrites ambiguous implicit user queries to prevent alignment errors in subsequent processing.

\textbf{GRPO Grounder Agent.} The GRPO Grounder Agent is designed to pinpoint the precise multi-modal elements that align with the user query. Inspired by its effectiveness in LLM training, we adopt GRPO to train the Grounder \cite{guo2025deepseek}.

\textbf{Omni-Modal Long-Tail Intention Understanding (OMLT) Reasoner Agent.} Acting as a System-2 module \cite{kahneman2011thinking}, the OMLT Reasoner Agent applies instance-specific knowledge distillation to deeply analyze long-tail events rich in human intent and emotion. Utilizing CoT reasoning, it step-by-step resolves the user query based on these omni-modal observations. 

\textbf{Test-Time Adaption (TTA) Reviser Agent.} TTA Reviser Agent assesses the OMLT Reasoner's output, initiating a closed feedback loop if the answer is suboptimal. Through Episodic LoRA Refinement, it dynamically updates the Reasoner's parameters during inference to regenerate the response. Importantly, once a satisfactory answer is produced, all LoRA-induced modifications are instantly discarded to protect the base model from catastrophic forgetting \cite{luo2025empirical}.

Social MLLMs remain a primary challenge, reliant on both intrinsic model capabilities and strategic agent-level orchestration. In MODF-SIR, we employ targeted prompting to force the MLLM to extract and linguistically amplify hidden long-tail events via detailed textual descriptions. This linguistic amplification prevents critical but subtle cues from being drowned out by dominant visual features or background noise during text tokenization. As a result, the LLM can reliably attend to these nuances, anchoring its extended CoT reasoning on these explicit social signals.

\textbf{Dynamic Cognitive Workflow.} 
Unlike unidirectional pipelines, MODF-SIR employs a dynamic, closed-loop workflow comprising three phases:
\textit{Phase 1 Global Context Distillation.} The ELT Retriever Agent extracts multi-modal cues from downsampled inputs to produce a concise semantic summary.
\textit{Phase 2 Dynamic Reasoning Routing.} The AKD Router Agent determines the processing path. Sufficient evidence takes Path A directly to the OMLT Reasoner. Complex queries take Path B, where the GRPO Grounder Agent performs precise spatio-temporal localization before reasoning to mitigate fragmentation errors.
\textit{Phase 3 Reflexive Generation \& Test-Time Adaptation.} The OMLT Reasoner Agent generates a candidate response and reasoning chain. The TTA Reviser Agent then self-evaluates, applying dynamic rewards for test-time adaptation to correct any logical inconsistencies or hallucinations.
\subsection{ELT Retriever Agent}
Multi-modal representation learning has shifted from coarse-grained alignment to fine-grained reasoning. Early contrastive models (e.g., CLIP \cite{radford2021learning}, HowTo100M \cite{miech2019howto100m}) excel at clip-level retrieval but use static vectors that miss fine-grained temporal and affective details in long videos. Subsequent methods introduced temporal grounding and dense fusion (e.g., TALL \cite{gao2017tall}, MERLOT \cite{zellers2021merlot}) for context-aware reasoning. However, relying on frozen feature extractors restricts their ability to generate interpretable, semantically rich natural language representations.
Retrieval-Augmented Generation (RAG) \cite{lewis2020retrieval} mitigates LLM hallucinations by retrieving external knowledge, a paradigm increasingly extended to multi-modal domains. However, conventional RAG relies on pre-segmented databases and shallow similarity metrics, limiting its effectiveness in long-form video reasoning. In such videos, implicit signals (e.g., intentions, emotional dynamics) are embedded within continuous streams, discrete chunking disrupts their temporal coherence. This necessitates a new paradigm: internal, generative information extraction directly from continuous audio-visual streams, replacing external, discrete feature matching.
To avoid the prohibitive costs and semantic noise of processing uncompressed audio-visual streams, the ELT Retriever Agent extracts query-relevant long-tail events (encapsulating subtle intents and emotions). It generates a natural language description, $C_{text}$, which guides the subsequent AKD Router Agent's decisions.
Given a raw video stream $V$ and its corresponding audio stream $A$, we first apply a downsampling operator $\mathcal{D}(\cdot)$ to extract keyframes and reduce spatio-temporal resolution, yielding compressed representations $V’ = \mathcal{D}(V)$ and $A’ = \mathcal{D}(A)$. The ELT Retriever Agent is instantiated using an omni-modal foundation model, such as Qwen2.5-Omni \cite{xu2025qwen3}. It takes as input the compressed streams $V’$ and $A’$, the user query $Q$, and a system prompt $P$. 

The objective of this module is to proactively scan the multimodal inputs, identify query-relevant cues, and synthesize them into a concise textual context $C_{\text{text}}$. Formally, this generative extraction process is defined as:
\begin{equation}
    C_{text} = \mathcal{M}_{retriever}(V', A', Q, P).
\end{equation}
Here, the foundational MLLM $\mathcal{M}_{retriever}$ powers our ELT Retriever Agent to directly output natural language ($C_{text}$), rather than rigid timestamps or vectors. This text explicitly captures query-relevant audio-visual cues as a semantic prior. It then directly guides the downstream AKD Router Agent's reasoning, effectively filtering out redundant backgrounds and reducing cognitive load.

\subsection{AKD Router Agent}
\textbf{C1 Routing Decision.} To address the trade-off between performance and efficiency, we draw inspiration from the intuitive, fast-thinking mechanism of human “System 1” \cite{kahneman2011thinking} and propose an Adaptive Reasoning AKD Router Agent. 
The AKD Router Agent dictates the subsequent processing pathway by evaluating two factors: whether the user query is explicit or implicit, and whether the ELT Retriever Agent's textual output, $C_{text}$, describes a head or long-tail event.
Specifically, an explicit query is defined as a direct and unambiguous request that provides a comprehensive description of the user's intent and clear target objectives. Conversely, an implicit query is characterized by indirect formulation, where the true underlying objective is obscured beneath the surface statement and must be inferred through contextual or situational cues.
\begin{figure}[htbp]
    \centering
    \includegraphics[trim=1mm 1mm 1mm 1mm, clip, width=\linewidth]{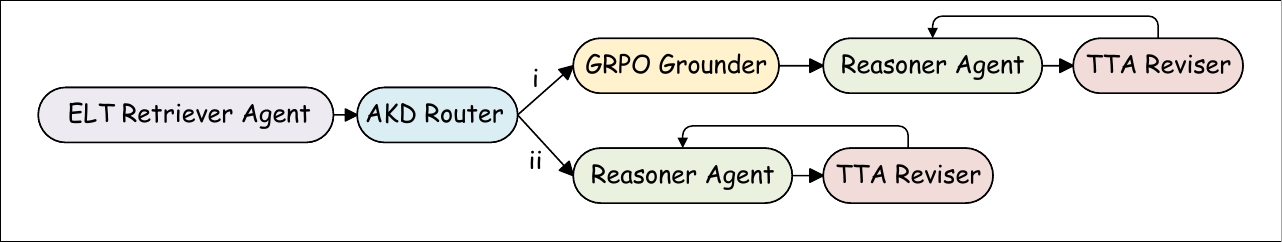}
    \caption{The AKD Router Agent activates other agents based on the video, audio and query, providing two strategies for different needs.}
    \label{planner}
\end{figure}
Formally, head events are defined as high-frequency occurrences within multi-modal data that exhibit significant cross-modal feature consistency \cite{geirhos2020shortcut}. MLLMs can accurately identify these events relying solely on superficial pattern matching, circumventing the need for deep causal reasoning. Conversely, long-tail events are infrequent occurrences characterized by subtle, and often contradictory, cross-modal signals \cite{thrush2022winoground}. Robust identification of these events forces models to move beyond statistical averages, demanding dynamic, fine-grained semantic grounding and logical inference. In essence, the multi-modal representations of subtle emotional expressions or sudden shifts in human intent fundamentally constitute long-tail events \cite{lian2024ov}.

When the AKD Router Agent identifies an explicit query accompanied predominantly by head events in $C_{text}$, it outputs a \texttt{[\{"type": "Answer"\}]} flag. This signals the framework to directly activate the OMLT Reasoner, enabling it to infer the response directly from the raw multi-modal data stream without requiring further complex routing.
If the AKD Router Agent detects an implicit query coupled predominantly with long-tail events in $C_{text}$, it generates the routing sequence \texttt{[\{"type": "Grounder", "value": "}$Q_{\text{ground}}$\texttt{"\}, \{"type": "Answer"\}]}. This triggers a mandatory grounding phase before final reasoning. The Router explicitly disambiguates the obscure query into a highly comprehensible format, $Q_{ground}$. Using this $Q_{ground}$, the subsequent GRPO Grounder performs spatiotemporal localization on the multimodal data, providing the foundation for the final reasoning stage.

\textbf{C2 Training via Asymmetric Knowledge Distillation and Cognitive Compression.} Despite the effectiveness of the proposed AKD Router Agent, training is hindered by a scarcity of supervision for multimodal routing of human intent. To address this limitation, we introduce an asymmetric knowledge distillation paradigm that enables cognitive compression, eliminating the need for costly and error-prone manual annotations. During data construction, we employ a large-scale omni-modal foundation model, such as Qwen3-Omni-30B-A3B-Thinking \cite{xu2025qwen3}, as the teacher. Given multimodal inputs rich in human intent, along with carefully designed expert prompts, the teacher performs unconstrained offline reasoning via CoT to generate pseudo-labels, including optimal routing decision trees and reformulated queries $Q_{\text{ground}}$, conditioned on query complexity. We then instantiate the AKD Router Agent using a lightweight student model, such as Qwen2.5-Omni-7B \cite{xu2025qwen3}. The student is trained via SFT with LoRA on the high-quality pseudo-labels produced by the teacher. This process effectively distills the teacher’s computationally intensive planning capability into a compact model, compressing the high-level strategic reasoning of a 30B-parameter model into a 7B-parameter network. As a result, the AKD Router Agent acquires a level of strategic competence that exceeds its parameter scale at inference time. It can efficiently orchestrate the multi-agent system with precise routing decisions while maintaining low computational overhead and latency.

\subsection{GRPO Grounder Agent}
The GRPO Grounder operates exclusively within Path B of the AKD Router's architecture. Routing to this path implies the query requires complex intent or emotion analysis derived from long-tail cues. Given the subtle and diverse nature of long-tail events, exhaustive global extraction across the unconstrained multi-modal data stream is computationally prohibitive \cite{zhang2023deep}. Consequently, the GRPO Grounder is designed to spatio temporally localize the most query-relevant data segments. This critical step restricts the search space for subsequent event extraction and extended reasoning to a computationally tractable scale.

Video remains the most prominent and widely used data modality. For video inputs, the GRPO Grounder's objective is to temporally localize the segment most relevant to the user query. During supervised training, the GRPO Grounder takes the raw video and query as inputs to predict a temporal boundary. The deviation between this predicted segment and the ground-truth annotation is used to construct a loss function. By backpropagating the resulting gradients, the GRPO Grounder's parameters are dynamically updated, iteratively refining the model to generate predictions that closely approximate the ground-truth boundaries.
Most existing video grounders are anchored by a Vision Transformer (ViT) backbone \cite{dosovitskiy2020image}. As a form of MLLM, these architectures combine video feature extractors with LLMs. Inspired by the performance gains achieved by GRPO \cite{shao2024deepseekmath} in LLM optimization, we adopt the GRPO paradigm to train and refine our GRPO Grounder agent.

To address the challenges of long-form video processing and the semantic ambiguity of user queries, the Grounder outputs a set of $N$ candidate segments $\{y_1, \cdots, y_N\}$ with associated confidence scores $\{c_1, \cdots, c_N\}$. The reward for each proposed segment is computed as its degree of overlap with the ground-truth segment $z$, providing a supervised signal for the grounding task:
\begin{equation}
    R(y_t) = \frac{\text{area}(z \cap y_t)}{\text{area}(z \cup y_t)} \quad \text{or} \quad R(y_t) = \frac{\text{area}(z \cap y_t)}{\text{area}(z)}.
\end{equation}
Here, $\text{area}(\bullet)$represents the duration of the video segment. Thus, we can calculate the relative rewards within the group of GRPO:
\begin{equation}
        \tilde{R}(y_t) = \frac{R(y_t) - \mu}{\sigma},
\end{equation}
where $\mu = \frac{1}{N} \sum_{t=1}^{N} R(y_t)$ and $\sigma = \sqrt{\frac{1}{N} \sum_{t=1}^{N} (R(y_t) - \mu)^2}$ represent group mean and group standard deviation respectively. Then we calculate:
\begin{equation}
    r(y|x; \theta) = \frac{\pi(y|x; \theta)}{\pi_{old}(y|x)} = \prod_{t=1}^{N} \frac{\pi(y_t|x; \theta)}{\pi_{old}(y_t|x)} \in \mathbb{R}.
\end{equation}
Here $\theta$ denotes the GRPO Grounder parameters and $x$ the input (video and query). The conditional probability of outputting segment $y$, denoted as $\pi_{\theta}(y|x)$, is computed via Softmax over the confidence scores $\{c_1, \dots, c_N\}$:
\begin{equation}
    \rho(\bullet | x; \theta) = \text{softmax}
    \begin{bmatrix}
        \hat{c}_{i_1} \\
        \vdots \\
        \hat{c}_{i_N}
    \end{bmatrix}
    =
    \begin{bmatrix}
        \frac{e^{\hat{c}_{i_1}}}{\sum_{t=1}^{N} e^{\hat{c}_{i_t}}} \\
        \vspace{-0.5em} \vdots \\
        \frac{e^{\hat{c}_{i_N}}}{\sum_{t=1}^{N} e^{\hat{c}_{i_t}}}
    \end{bmatrix}
    \in \mathbb{R}^N,
\end{equation}
\begin{equation}
    \pi(y | x; \theta) = \prod_{t=1}^{N} \rho(y_t | x; \theta) \in \mathbb{R}.
\end{equation}

Within the iterative training process, $\pi_{old}$ represents the model state from the preceding step, with the current $\pi_{\theta}$ becoming the new $\pi_{old}$ for the next iteration. Accordingly, the loss function for the GRPO Grounder is defined as:
\begin{multline}
    L_{act}(x, \theta) = \frac{1}{N} \sum_{t=1}^{N} \min \{ r(y_t, x, \theta)\tilde{R}(y_t), \\
    \text{clip}(r(y_t, x, \theta), 1-\varepsilon, 1+\varepsilon)\tilde{R}(y_t) \},
\end{multline}

\begin{equation}
    L_{\text{GRPO}}(\theta) = L_{act}(x, \theta) + \beta \text{KL}(\pi \parallel \pi_{old}; x, \theta).
\end{equation}
Here, the clipping function and the KL divergence are defined as:
\begin{multline}
    \text{clip}(r, 1-\varepsilon, 1+\varepsilon) = 
    \begin{cases} 
        r, & 1-\varepsilon \le r \le 1+\varepsilon \\ 
        1-\varepsilon, & r < 1-\varepsilon \\ 
        1+\varepsilon, & r > 1+\varepsilon 
    \end{cases} \\
    = \min(\max(r, 1-\varepsilon), 1+\varepsilon),
\end{multline}

\begin{equation}
    \text{KL}(\pi \parallel \pi_{old}; x, \theta) = \sum_{t=1}^{N} \pi(y_t|x; \theta) \log \frac{\pi(y_t|x; \theta)}{\pi_{old}(y_t|x)}.
\end{equation}
By optimizing the GRPO objective, the Grounder learns to perform precise continuous-time localization that is directly aligned with the global IoU metric. By constraining the reasoning process to these grounded segments, the framework effectively suppresses irrelevant context and mitigates shortcut-induced hallucinations.


\begin{figure*}[htbp]
    \centering
    \includegraphics[trim=1mm 1mm 1mm 1mm, clip, width=\linewidth]{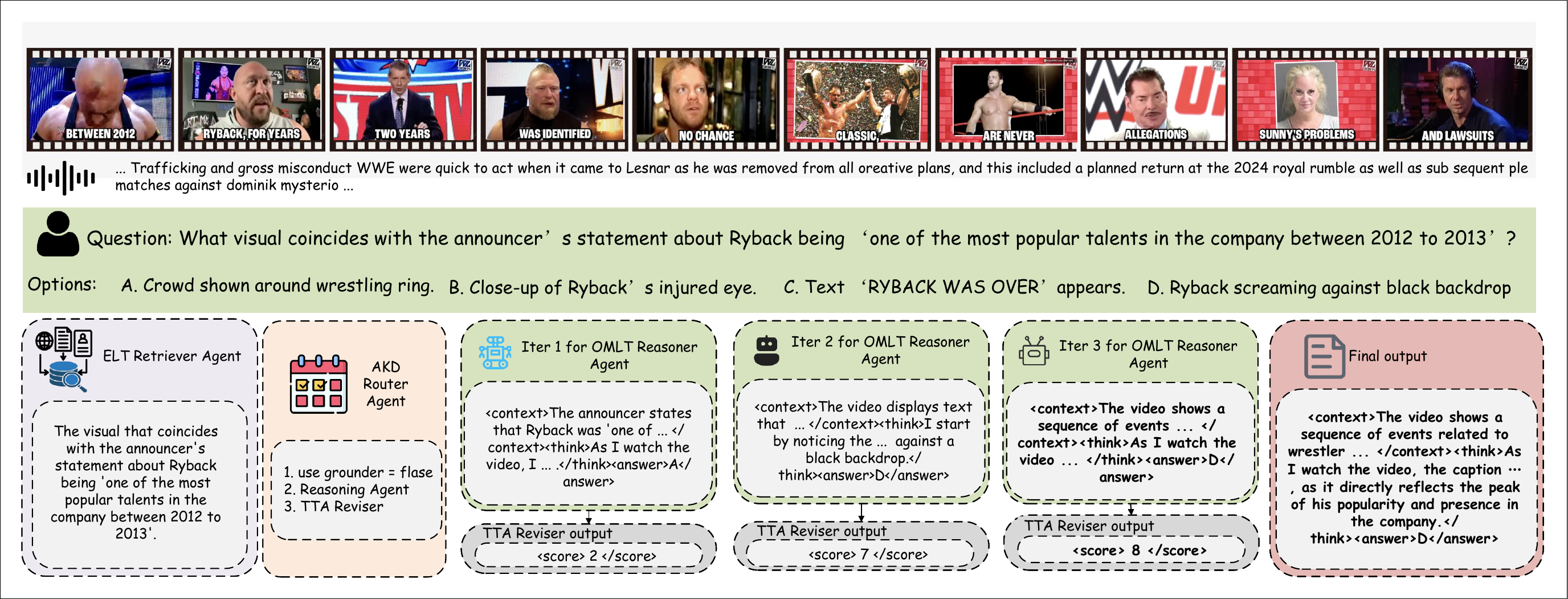}
    \caption{\textbf{Visualization of our MODF-SIR.} ELT Retriever Agent would first collect the information from the video, audio and query. The AKD Router Agent would first determine which mode to use. In this sample, AKD Router Agent decide to use OMLT Reasoner Agent directly. OMLT Reasoner Agent generates responses, and TTA Reviser scores these responses. If the score is below the satisfaction threshold, LoRA fine-tuning is performed on OMLT Reasoner Agent, and then the answers are regenerated.}
    \label{visualization}
\end{figure*}

\subsection{OMLT Reasoner Agent}
In social intelligence reasoning, true intentions are often revealed not by explicit speech, but by subtle, long-tail micro-events (e.g., micro-expressions, vocal tremors) \cite{schuller2022acm, ekman2003darwin, goyal2017something}. However, modeling these presents two challenges: (i) Sparse Signal Concentration, due to an extremely low signal-to-noise ratio in high-dimensional omni-modal streams \cite{feichtenhofer2019slowfast} and (ii) Implicit Causal Coupling, which requires integrating asynchronous, sometimes contradictory multi-modal cues to resolve ambiguity \cite{tsai2019multimodal}.

To navigate this complex, open-ended problem space, we propose the \textbf{OMLT Reasoner Agent}. 
Anchored by a lightweight MLLM, the OMLT Reasoner Agent utilizes extended CoT reasoning to deduce the answer to the user's query. This inference is systematically guided by the input multi-modal data, the original query, and the AKD Router Agent's control sequence.
If the AKD Router Agent outputs the \texttt{[\{"type": "Answer"\}]} control sequence, the OMLT Reasoner Agent directly infer the response from the raw multi-modal data. Conversely, if the Router generates \texttt{[\{"type": "Grounder", "value": "}$Q_{\text{ground}}$\texttt{"\}, \{"type": "Answer"\}]}, it signifies that the user's input is an implicit query demanding human intent or emotional dynamics. In this scenario, the OMLT Reasoner must first extract the nuanced long-tail events from the grounded multi-modal data. It executes CoT reasoning to systematically deduce the final answer. To achieve this, the operational workflow of the OMLT Reasoner proceeds as follows:
(i) OMLT Reasoner Agent dynamically extracts query-relevant and long-tail events from the multi-modal stream, formulating a textual description termed Context. 

Although ostensibly similar to the ELT Retriever, the two modules have fundamentally differences.
First, their inputs differ significantly. The ELT Retriever ingests raw audio-visual data and the original user query. Conversely, the OMLT Reasoner operates on the AKD Router's disambiguated query  and the GRPO Grounder's precisely localized video.
Second, their operational objectives differ. The ELT Retriever acts solely as an objective evidence gatherer to inform the AKD Router. 
Conversely, the OMLT Reasoner operates on the premise of social intelligence analysis, meticulously examining the grounded video to extract query-specific long-tail events. Grounded in ``Dual-Process'' Theory, the Retriever performs System-1 heuristic scanning, whereas the Reasoner executes System-2 analytical processing.
(ii) The OMLT Reasoner submits Context, $V_{ground}$, and $Q_{ground}$ to a powerful external MLLM for evaluation. If the assigned score is below a predefined threshold, the Reasoner's base model applies LoRA to regenerate the Context. This process acts as an iterative Teacher-Student knowledge distillation loop. The external MLLM (Teacher) critiques the output of the base model (Student), enforcing iterative LoRA updates until the generated Context satisfies the required quality criteria.
(iii) Upon approval of the Context, the OMLT Reasoner executes extended CoT reasoning based on Context and $Q_{ground}$ to generate the reasoning trace Think, the conclusion of which serves as the answer. Consistent with the context validation step, the Reasoner submits Think, Context, and $Q_{ground}$ to an external LLM for scoring. Scores below a predefined threshold trigger LoRA-based fine-tuning of the base model to regenerate Think. This teacher-student distillation process culminates in the final answer, which is output to the TTA Reviser.

Crucially, the OMLT Reasoner's parameter updates are subject to two constraints. First, the LoRA fine-tuning is an additive adjustment, formulated as $W = W_0 + \Delta W$, where $W_0$ represents the original base parameters and $\Delta W$ is the adaptive update. Second, this adaptation is entirely instance-specific. Once a satisfactory inference is achieved for the current video and query, the $\Delta W$ weights are discarded, reverting the base model to its initial state to completely prevent catastrophic forgetting.

\subsection{TTA Reviser Agent}
\textbf{F1 Test-Time Adaption.} 
The TTA Reviser evaluates the final answer generated by the OMLT Reasoner. Its integration is driven by the well-established generation-evaluation gap, which posits that an LLM's capacity to assess a result inherently surpasses its ability to generate one \cite{lightman2023let}.
The TTA Reviser and OMLT Reasoner share a single base model. While the Reasoner uses this foundation for generation, the Reviser leverages the exact same model for evaluation. Rooted in the generation-evaluation gap, this shared-architecture approach yields a highly robust and credible assessment.
Using targeted prompting, the TTA Reviser tasks the shared base model with scoring the OMLT Reasoner's output. Should the score fall below a set threshold, the Reviser employs TTA via LoRA fine-tuning to dynamically adjust the Reasoner and yield an improved answer.
The OMLT Reasoner updates the base model via LoRA to generate an initial answer, expressed as:
\begin{equation}
    W_{reason}=W_0+\Delta\phi, \quad \Delta\phi=BA,
\end{equation}
where $W_0$ represents the base parameters, and $A$ and $B$ are optimized low-rank matrices. 
To refine this output, the TTA Reviser sequentially applies another LoRA-based update to the OMLT Reasoner's modified weights:
\begin{equation}
    W_{reviser}=W_{reason}+\Delta\varphi, \quad \Delta\varphi=CD.
\end{equation}
Here, optimizing the secondary low-rank matrices $C$ and $D$ enables the TTA Reviser to dynamically generate an improved answer.
This procedure functions as an iterative feedback loop. The TTA Reviser continuously evaluates the OMLT Reasoner's output, enforcing iterative regeneration until a quality threshold is met or a maximum iteration limit is reached. Post-inference, the learned incremental weights ($\Delta\phi$ and $\Delta\varphi$) are instantly discarded, restoring the parameters $W_{reviser}$ directly to the original base weights $W_0$. This ephemeral parameter update paradigm formally constitutes our TTA approach.

\textbf{F2 Single sample optimizations Reinforce algorithm.} 
In contrast to the GRPO Grounder, which relies on multiple samples for relative reward computation, the TTA Reviser processes only a single output from the OMLT Reasoner, rendering GRPO inapplicable. Moreover, the Reasoner's discrete text output is non-differentiable. To resolve this, we model the text generation process as a Markov Decision Process (MDP). This allows us to treat the Reviser's evaluation as a parameter-independent reward and apply the REINFORCE algorithm, which is ideal for single-sample optimization, to effectively fine-tune the Reasoner’s parameters. The objective function for single sample optimizations Reinforce algorithm is as follows:
\begin{multline}
    \mathcal{J}(\Delta\phi) = \mathbb{E}_{y \sim \pi_{\theta + \Delta\phi}(\cdot | x)} \Big[ r_{\psi}(x, y) \\
    - \beta \mathbb{D}_{KL} \big(\pi_{\theta + \Delta\phi}(\cdot | x) \parallel \pi_{\theta}(\cdot | x) \big) \Big].
\end{multline}
Here, $x$ represents the OMLT Reasoner's input (video, audio and query), and $y$ represents its output answer, which is subsequently evaluated by the TTA Reviser. The reward $r_\psi(x, y)$ is the TTA Reviser's score, generated by the base model via targeted prompting. Let $\theta$ denote the OMLT Reasoner's original parameters, $\pi_{\theta+\Delta\phi}(\cdot|x)$ and $\pi_\theta(\cdot|x)$ then define the output conditional probabilities post and pre-update, respectively. $D_{KL}$ denotes the KL regularization term. 

Though formally identical to the KL penalty in GRPO, its objective here differs: within our REINFORCE setup, it mitigates large training oscillations induced by high iteration variance, whereas in GRPO, it serves to prevent reward hacking \cite{amodei2016concrete}.
Furthermore, by using the Log-Derivative Trick, the gradient of $\mathcal{J}(\Delta\phi)$can be obtained:
\begin{multline}
    \nabla_{\Delta\phi} \mathcal{J}(\Delta\phi) = \mathbb{E}_{y \sim \pi_{\theta + \Delta\phi}} [\nabla_{\Delta\phi} \log \pi_{\theta + \Delta\phi}(y|x) \\
    (\tilde{r}(x, y) - b_t) ],
\end{multline}
\begin{equation}
    \tilde{r}(x, y) = r_{\psi}(x, y) - \beta \log \frac{\pi_{\theta + \Delta\phi}(y|x)}{\pi_{\theta}(y|x)}.
\end{equation}
Here $b_t$ serves as an adaptive baseline employed for variance reduction.

When tackling exceptionally complex spatio-temporal reasoning, the TTA Reviser adheres to a rigorous exploration-exploitation and backtracking mechanism: \textbf{(i) Action Sampling and Evaluation.} The policy network samples a candidate response $y_t$, which is scored by the isomorphic evaluator as $r_t = r_{\psi}(x, y_t)$. If $r_t \ge \tau$, where $\tau$ denotes a predefined satisfaction threshold, the iteration terminates, indicating convergence. \textbf{(ii) Advantage-Driven Update.} If the response is suboptimal, the advantage is computed as $A_t = \tilde{r}(x, y_t) - b_{t-1}$. The transient parameters $\Delta\phi$ are then updated via a single-step gradient ascent using this advantage signal. \textbf{(iii) Baseline Smoothing.} To reduce variance under limited sampling, the baseline is updated using an exponential moving average (EMA): $b_t = \alpha b_{t-1} + (1 - \alpha)\tilde{r}(x, y_t)$, where $\alpha$ means attenuation rate of change. \textbf{(iv) Fail-Safe Backtracking.} A maximum iteration budget $T_{\max}$ is imposed. If the threshold $\tau$ is not reached, the system backtracks and selects the best response $y^* = \arg\max_{y \in \mathcal{H}} r_{\psi}(x, y)$ from the trajectory set $\mathcal{H} = \{y_1, \dots, y_{T_{\max}}\}$, ensuring robustness and a bounded performance guarantee.

\section{Experiments}
\subsection{Main Results}
In Tab. \ref{tab:daily-omni}, we compare our model on the Daily-Omni \cite{zhou2025daily} across multiple dimensions. \textbf{MODF-SIR} achieves an overall average of \textbf{64.9\%}, outperforming all existing open-source video-audio MLLMs and significantly narrowing the gap with proprietary systems. Compared with prior methods such as Unified-IO-2 \cite{lu2024unified}, VideoLLaMA2 \cite{cheng2024videollama}, and Qwen2.5-Omni \cite{xu2025qwen3}, \textbf{MODF-SIR} shows consistent and substantial improvements across all metrics, highlighting the robustness of our unified multi-modal modeling framework. 
Despite using a comparable model size of 7B, \textbf{MODF-SIR} approaches proprietary systems such as Gemini 2.0 Flash, highlighting its competitive efficiency and scalability. Overall, \textbf{MODF-SIR} establishes a new state-of-the-art among open-source video-audio MLLMs on Daily-Omni \cite{zhou2025daily}, demonstrating superior multi-modal perception, reasoning, and temporal understanding capabilities.
  \begin{table*}[t!]
  \centering
  \caption{Comparison with other methods on Daily-Omni \cite{zhou2025daily}}
  \label{tab:daily-omni}
  \resizebox{\textwidth}{!}{%
  \begin{tabular}{lc|cccccc|cc|c}
  \hline
  Methods & \begin{tabular}[c]{@{}c@{}}LLM\\ Size\end{tabular} & \begin{tabular}[c]{@{}c@{}}AV Event\\ Alignment\end{tabular} & Comparative & \begin{tabular}[c]{@{}c@{}}Context\\ Understanding\end{tabular} & \begin{tabular}[c]{@{}c@{}}Event\\ Sequence\end{tabular} & Inference & Reasoning & \begin{tabular}[c]{@{}c@{}}30s\\ Subset\end{tabular} & \begin{tabular}[c]{@{}c@{}}60s\\ Subset\end{tabular} & Avg \\ \hline
  \multicolumn{11}{c}{\textit{\textbf{Proprietary MLLMs}}} \\ \hline
  Gemini 2.0 Flash  & - & 62.2 & 73.3 & 63.7 & 63.7 & 76.6 & 75.4 & 62.3 & 56.6 & 67.8 \\
  Gemini 2.0 Flash Lite  & - & 55.0 & 64.9 & 58.0 & 54.3 & 74.0 & 72.0 & 60.6 & 53.0 & 61.3 \\ \hline
  \multicolumn{11}{c}{\textit{\textbf{Open-Source Video-Audio MLLMs}}} \\ \hline
  Unified-IO-2 \cite{lu2024unified}    & 8B & 25.6 & 31.3 & 26.4 & 25.8 & 35.1 & 29.7 & 26.7 & 30.0 & 28.2 \\
  VideoLLaMA2 \cite{cheng2024videollama}   & 7B & 35.7 & 35.9 & 35.8 & 31.7 & 40.9 & 34.3 & 38.0 & 31.8 & 35.2 \\
  Qwen2.5-Omni \cite{xu2025qwen3}  & 3B & 38.7 & 48.1 & 33.7 & 34.0 & 54.56 & 44.0 & 46.7 & 48.4 & 40.5 \\
  Qwen2.5-Omini \cite{xu2025qwen3}  & 7B & 44.1 & 51.2 & 38.9 & 40.5 & 57.8 & 61.7 & 42.4 & 38.4 & 47.5 \\
  Ola \cite{liu2025ola} & 7B & 40.3 & 60.3 & 39.9 & 44.1 & 61.0 & 66.3 & 50.9 & 48.7 & 49.9 \\
  MiniCPM-o \cite{liu2025ola}  & 7B & 40.3 & 61.1 & 49.2 & 48.4 & 68.8 & 61.1 & 54.1 & 52.0 & 53.1 \\
  HumanOmniV2 \cite{yang2025humanomniv2}  & 7B & 46.6 & 67.9 & 51.8 & 51.6 & 72.7 & 74.3 & 63.1 & 53.1 & 58.5 \\
  \rowcolor[HTML]{D6D2D2}
  \textbf{MODF-SIR (Ours)} & 7B & \textbf{56.4} & \textbf{73.2} & \textbf{61.3} & \textbf{57.2} & \textbf{75.9} & \textbf{78.2} & \textbf{67.2} & \textbf{61.7} & \textbf{64.9} \\ \hline
  \end{tabular}
  }
\end{table*}

\begin{figure*}[t]
    \centering
    \begin{minipage}[c]{0.65\textwidth}
        \centering
        \captionof{table}{Comparison with other methods on IntentBench \cite{yang2025humanomniv2}}
        \label{tab:comparison_intentbench}
        \small 
        \begin{tabular}{cc|ccccc|c} 
        \hline
        \multirow{2}{*}{Methods} & \multirow{2}{*}{\begin{tabular}[c]{@{}c@{}}LLM\\ Size\end{tabular}} & \multicolumn{5}{c|}{Social} & \multirow{2}{*}{Avg} \\  &  & Why & How & When & Who/Which & Other &  \\ \hline
        \multicolumn{8}{c}{\textit{\textbf{Proprietary MLLMs}}} \\ \hline
        GPT-4o\cite{hurst2024gpt}  & -  & 61.5 & 55.7 & 35.7 & 76.0 & 63.3 & 60.0 \\
        GPT-o1 \cite{jaech2024openai} & -   & 68.2 & 65.8 & 57.1 & 76.0 & 68.3 & 66.7 \\
        Gemini-2.5-Pro (think)\cite{google2025gemini25pro}   & - & 68.6 & 67.4 & 57.1 & 64.0 & 70.0 &  67.2 \\ \hline
        \multicolumn{8}{c}{\textit{\textbf{Open-Source Video-Audio MLLMs}}} \\ \hline
        MiniCPM-o \cite{yao2024minicpm}   & 8B   & 57.9 & 53.5 & 57.1 & 68.0 & 61.1 & 54.5 \\
        VITA-1.5  \cite{fu2025vita}   & 7B & 53.2 & 49.4 & 71.4 & 64.0 & 61.1 &  54.2 \\
        Ola   \cite{liu2025ola} & 7B & 60.6 & 55.4 & 64.3 & 76.0 & 62.9 & 57.4 \\
        Qwen2.5-Omni  \cite{xu2025qwen3}   & 7B  & 62.6 & 63.4 & 57.1 & 76.0 & 69.0 &  64.2 \\
        HumanOmniV2   \cite{yang2025humanomniv2}   & 7B  & 66.8 & 67.1 & 50.0 & 84.0 & 72.4 &  69.3 \\ \hline
        \rowcolor[HTML]{D6D2D2} 
        \textbf{MODF-SIR  (Ours)} & \textbf{7B}  & \textbf{67.2} & \textbf{69.5} & \textbf{64.3} & \textbf{84.0} & \textbf{75.5} &  \textbf{70.3} \\ \hline
        \end{tabular}
    \end{minipage}%
    \hfill 
    \begin{minipage}[c]{0.3\textwidth}
        \centering
        \includegraphics[trim=1mm 1mm 1mm 1mm, clip, width=\linewidth]{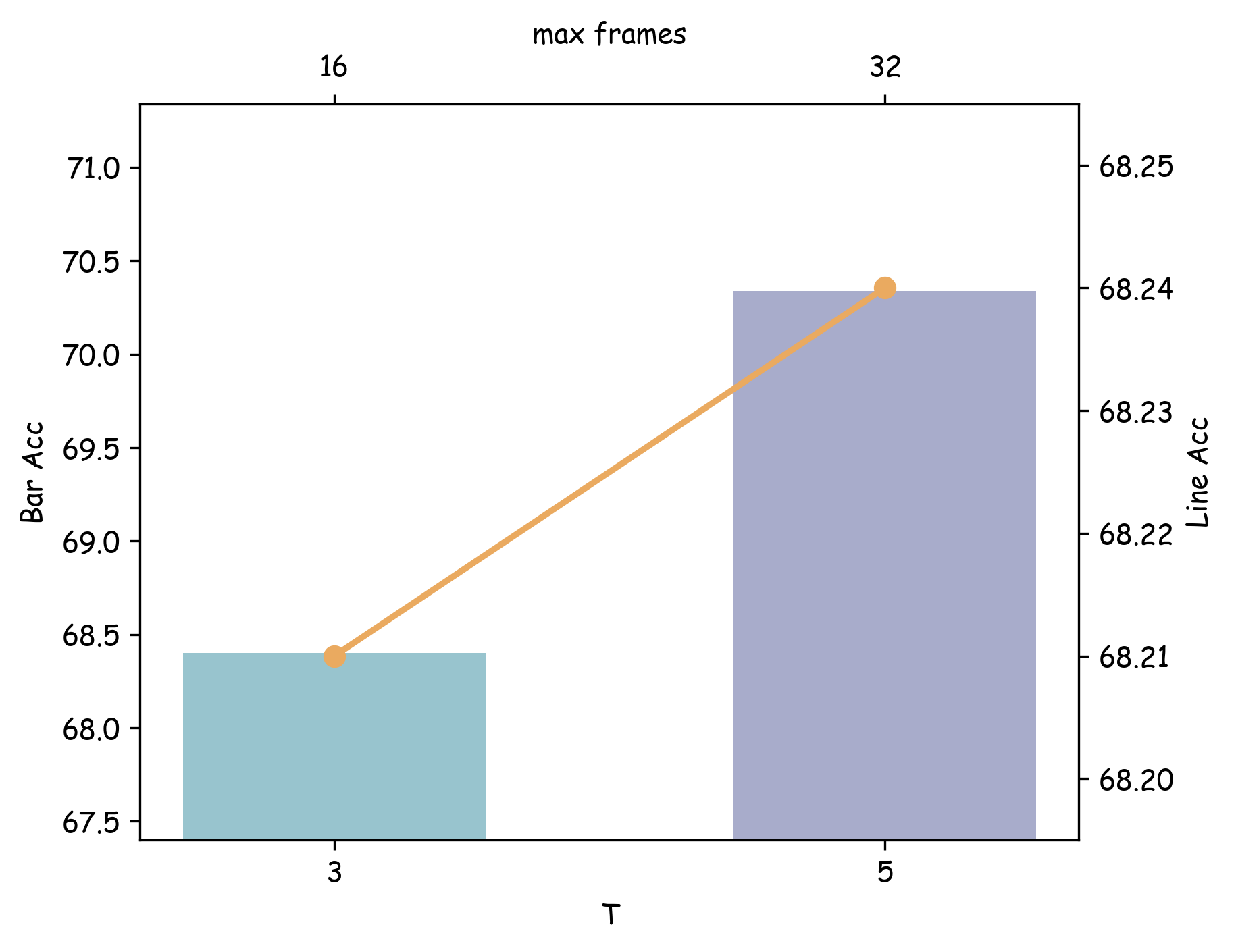}
        \captionsetup{font=scriptsize}
        \captionof{figure}{T means the TTA Reviser’s maximum iterations (left axis, bars), while ``max frames'' means the OMLT Reasoner Agent’s maximum input frames (right axis, line).}
        \label{bar_ib}
    \end{minipage}
\end{figure*}

\begin{table*}[t]
  \centering
  \caption{Comparison with other methods on WorldSense \cite{hong2025worldsense}}
  \label{tab:comparison_worldsense}
  \resizebox{\textwidth}{!}{%
  \begin{tabular}{cc|cccccccc|c}
  \hline
  Methods & \begin{tabular}[c]{@{}c@{}}LLM\\ Size\end{tabular} & \begin{tabular}[c]{@{}c@{}}Tech\&\\ Science\end{tabular} & \begin{tabular}[c]{@{}c@{}}Culture\&\\ Politics\end{tabular} & \begin{tabular}[c]{@{}c@{}}Daily\\ Life\end{tabular} & \begin{tabular}[c]{@{}c@{}}Film\&\\ TV\end{tabular} & Performance & Games & Sports & Music & Avg \\ \hline
  \multicolumn{11}{c}{\textit{\textbf{Proprietary MLLMs}}} \\ \hline
  Claude 3.5 Sonnet \cite{anthropic2024introducing} & - & 43.7 & 31.7 & 30.6 & 36.5 & 30.7 & 31.9 & 36.6 & 33.9 & 34.8 \\
  GPT 4o \cite{hurst2024gpt} & - & 48.0 & 44.0 & 38.3 & 43.5 & 41.9 & 41.2 & 42.6 & 42.6 & 42.6 \\
  Gemini 1.5 Pro \cite{team2024gemini} & - & 53.7 & 47.2 & 50.3 & 5.4 & 52.4 & 46.8 & 40.2 & 42.0 & 48.0 \\ \hline
  \multicolumn{11}{c}{\textit{\textbf{Open-Source Video-Audio MLLMs}}} \\ \hline
  Unified-IO-2 XXL \cite{lu2024unified} & 7B & 27.1 & 31.7 & 23.9 & 23.7 & 25.5 & 23.7 & 25.7 & 27.3 & 25.9 \\
  VideoLLaMA2 \cite{cheng2024videollama} & 7B & 29.4 & 25.4 & 21.8 & 24.5 & 26.2 & 24.6 & 25.5 & 27.1 & 25.4 \\
  VITA-1.5 \cite{fu2025vita} & 7B & 38.2 & 35.9 & 34.3 & 39.8 & 41.2 & 32.6 & 34.7 & 39.9 & 36.9 \\
  Qwen2.5-Omini \cite{xu2025qwen3}& 7B & 47.8 & 49.8 & 43.6 & 43.8 & 48.3 & 39.1 & 43.5 & 47.3 & 45.4 \\
  HumanOmniV2 \cite{yang2025humanomniv2} & 7B & 50.2 & 51.7 & 47.6 & 44.8 & 47.3 & 44.3 & 45.2 & 44.2 & 47.1 \\ \hline
  \rowcolor[HTML]{D6D2D2} 
  \textbf{MODF-SIR (Ours)} & \textbf{7B} & \textbf{54.7} & \textbf{58.2} & \textbf{50.3} & \textbf{49.6} & \textbf{52.1} & \textbf{51.5} & \textbf{47.7} & \textbf{50.0} & \textbf{51.5} \\ \hline
  \end{tabular}
  }
\end{table*}

\begin{table}[]
\caption{Ablation results on our method.}
\fontsize{6.8pt}{7.5pt}\selectfont
\begin{tabular}{l|l}
\hline
Method  & Avg   \\ \hline
HumanomniV2 (Baseline) \cite{yang2025humanomniv2}   & 58.5 \\
\textbf{Our (Full)}    & \textbf{64.9} \\ \hline
\textbf{OMLT Reasoner Agent + TTA Reviser}  & \textbf{64.0} \\
\textbf{OMLT Reasoner Agent + TTA Reviser + GRPO Grounder} & \textbf{58.6} \\
OMLT Reasoner Agent + TTA Reviser + VideoMind Grounder & 57.3 \\
OMLT Reasoner Agent + TTA Reviser + GRPO Grounder + AKD Router & 59.4 \\ \hline
\end{tabular}
\label{ablation}
\end{table}

We report results on IntentBench \cite{yang2025humanomniv2} in Tab. \ref{tab:comparison_intentbench}. \textbf{MODF-SIR} achieves an average score of \textbf{70.3\%}, outperforming the strongest open-source baseline, HumanOmniV2 \cite{yang2025humanomniv2}, and remaining competitive with proprietary models such as GPT-o1 \cite{jaech2024openai} and Gemini-2.5-Pro \cite{google2025gemini25pro}. This demonstrates the strong capability of \textbf{MODF-SIR} in human intentions understanding under a comparable model scale. 
These improvements suggest that \textbf{MODF-SIR} is more effective in capturing both explicit intent signals and implicit contextual cues, enabling more precise and reliable user-centric multi-modal reasoning. Furthermore, as illustrated in Fig. \ref{bar_ib}, increasing the number of iterations $T$ and max frames consistently improves performance on IntentBench \cite{yang2025humanomniv2}.
In Tab. \ref{tab:comparison_worldsense}, we evaluate our model on WorldSense \cite{hong2025worldsense}, which assesses world knowledge and domain-specific understanding across diverse categories. \textbf{MODF-SIR} achieves an average score of \textbf{51.5\%}, significantly outperforming the strongest open-source baseline, HumanOmniV2 (47.1\%) \cite{yang2025humanomniv2}, while also approaching the performance of proprietary models. This demonstrates the effectiveness of \textbf{MODF-SIR} in modeling knowledge-intensive multimodal scenarios under a comparable model scale. 

\subsection{Ablation Studies}
\textbf{Effect of Individual Agents.} We conduct ablation studies on Daily-Omni \cite{zhou2025daily} (Tab. \ref{ablation}) to evaluate each module: \textbf{(i) TTA Reviser Agent:} It significantly improves the HumanOmniV2 \cite{yang2025humanomniv2} baseline from 58.5\% to 64.0\%. By utilizing isomorphic self-evaluation and ephemeral LoRA updates, the TTA Reviser Agent enables dynamic self-reflection, overcoming the limitations of conventional one-pass inference. \textbf{(ii) AKD Router Agent:} Adding the GRPO Grounder without the AKD Router Agent drops performance to 58.6\%, as indiscriminately applying temporal grounding introduces unnecessary visual noise. Reintroducing the AKD Router Agent restores performance to 59.4\%, proving the necessity of dynamic routing. \textbf{(iii) ELT Retriever Agent :} Comparing the 59.4\% configuration to the full model (64.9\%) highlights the ELT Retriever Agent's role. Without this initial abstraction module, the AKD Router Agent struggles with raw, uncompressed audio-visual inputs. \textbf{(iv) GRPO Grounder:} Although it causes noise in simple queries without the AKD Router Agent, it is crucial for precise evidence localization in complex, long-horizon queries. Properly integrated within the dynamic routing framework, it enables the system to achieve its peak score of 64.9\%.

\textbf{Effect of GRPO for Grounder.} To explicitly evaluate the proposed grounding optimization strategy, we compare the GRPO based Grounder with the conventional VideoMind Grounder under the same rigid routing setting. As shown in Table \ref{ablation}, the GRPO Grounder achieves an average score of 58.6\%, outperforming the VideoMind Grounder (57.3\%). This improvement stems from the difference in optimization objectives. Conventional grounding methods rely on local, pixel-level supervision, which is misaligned with the global temporal boundary metric. In contrast, our approach directly optimizes the non-differentiable IoU reward via GRPO, enabling better alignment with the evaluation criterion. These results demonstrate that GRPO facilitates more accurate continuous-time localization, producing higher-fidelity spatio-temporal segments that more effectively support downstream reasoning.

\section{Conclusions}
In this paper, we introduced  MODF-SIR, a pioneering multi-agent collaborative framework explicitly designed for omni-modal social intelligence reasoning. Comprehensive evaluations across the Daily-Omni, IntentBench, and WorldSense benchmarks demonstrate that our framework establishes new state-of-the-art standards for open-source models.
Establishing cross-temporal causal links between implicit long-tail events and preceding head events remains a core challenge in social intelligence reasoning. To address this, our future work proposes constructing a dedicated video shot segmentation agent, leveraging extensive computer vision research in shot boundary detection. This agent will partition videos into discrete, semantically stable segments, enabling an MLLM to generate localized textual descriptions. Subsequently, a LLM will be utilized to mine semantic correlations across these transcripts, effectively bridging the temporal gap between isolated shots to achieve comprehensive, cross-temporal reasoning.


\bibliographystyle{IEEEtran} 
\bibliography{ref}

\end{document}